%% file: main.tex
\documentclass[runningheads]{llncs}
\usepackage{graphicx}

\usepackage{tikz}
\usepackage{comment}
\usepackage{dsfont}
\usepackage{amsmath,amssymb} 
\usepackage{color, colortbl}
\usepackage{multirow}
\usepackage{threeparttable}
\usepackage{tabularx}
\usepackage{caption}
\usepackage{subcaption}
\usepackage{wrapfig}
\usepackage{hhline}
\usepackage{sidecap}
\usepackage[sort,nocompress]{cite}

\newcolumntype{C}{>{\centering\arraybackslash}X}
\newcolumntype{G}{>{\columncolor{yellow!30}\centering\arraybackslash}X}
\definecolor{VisRed}{rgb}{0.945, 0.105, 0.027}
\definecolor{VisOrange}{rgb}{0.999, 0.644, 0.000}
\definecolor{VisBlue}{rgb}{0.000, 0.000, 0.999}

\usepackage{xurl}

\begin{document}
\pagestyle{headings}
\mainmatter
\def\ECCVSubNumber{1356}  

\title{Prediction-Guided Distillation for Dense Object Detection} 

\titlerunning{Prediction-Guided Distillation}
%
\author{Chenhongyi Yang\inst{1}\thanks{Corresponding Author. Email: chenhongyi.yang@ed.ac.uk} \and
Mateusz Ochal\inst{2,3} \and
Amos Storkey\inst{2} \and
Elliot J. Crowley\inst{1}\index{Crowley, Elliot J.}}
\authorrunning{C. Yang et al.}
%
\institute{School of Engineering, University of Edinburgh, UK \and
School of Informatics, University of Edinburgh, UK \and
School of Engineering and Physical Sciences, Heriot-Watt University, UK
}
\input{term_define}
\maketitle

\input{sections/abstract}
\input{sections/intoduction}

\input{sections/related_works}
\input{sections/method_v2}

\input{sections/experiments}
\input{sections/conclusion}

\clearpage

{
\small
\bibliographystyle{splncs04}
\bibliography{kdbib}
}
\clearpage
\appendix

\input{sections/supp}

\end{document}

%% file: term_define.tex
\newcommand{\ourmethodIntro}{Prediction-Guided Distillation (PGD)} 
\newcommand{\ourmethod}{PGD} 
\newcommand{\ourmethodFull}{Prediction-Guided Distillation} 
\newcommand{\disregionIntro}{\emph{key predictive regions}} 
\newcommand{\disregion}{\emph{key predictive regions}} 
\newcommand{\fdrIntro}{Foreground Distillation Region (FDR)} 
\newcommand{\GT}{GT} 
\newcommand{\mateusz}[1]{\textcolor{blue}{\textbf{mateusz: }#1}}
\newcommand{\ourmoduleIntro}{\emph{Prediction-Guided Weighting} (PGW)}
\newcommand{\ourmodule}{PGW}

%% file: sections/abstract.tex
\begin{abstract}
Real-world object detection models should be cheap and accurate. Knowledge distillation (KD) can boost the accuracy of a small, cheap detection model by leveraging useful information from a larger teacher model. However, a key challenge is identifying the most informative features produced by the teacher for distillation. In this work, we show that only a very small fraction of features within a ground-truth bounding box are responsible for a teacher's high detection performance. Based on this, we propose~\ourmethodIntro, which focuses distillation on these~\emph{key predictive regions} of the teacher and yields considerable gains in performance over many existing KD baselines. In addition, we propose an adaptive weighting scheme over the key regions to smooth out their influence and achieve even better performance. Our proposed approach outperforms current state-of-the-art KD baselines on a variety of advanced one-stage detection architectures. Specifically, on the COCO dataset, our method achieves between +3.1\% and +4.6\% AP improvement using ResNet-101 and ResNet-50 as the teacher and student backbones, respectively. On the CrowdHuman dataset, we achieve +3.2\% and +2.0\% improvements in MR and AP, also using these backbones. Our code is available at \url{https://github.com/ChenhongyiYang/PGD}.

\keywords{Dense Object Detection, Knowledge Distillation}
\end{abstract}


%% file: sections/intoduction.tex
\section{Introduction}

Advances in deep learning have led to considerable performance gains on object detection tasks~\cite{cai2018cascade,chen2021ddod,he2017mask,li2020generalized,lin2017focal,RedmonDiGiFa16,redmon2017yolo9000,ren2015faster,tian2019fcos}. However, detectors can be computationally expensive, making it challenging to deploy them on devices with limited resources. 
Knowledge distillation (KD)~\cite{ba2014do,hinton2015distilling} has emerged as a promising approach for compressing models. It allows for the direct training of a smaller student model~\cite{li2017light,qin2019thundernet,sandler2018mobilenetv2,wang2018pelee} using information from a larger, more powerful teacher model; this helps the student to generalise better than if trained alone.

KD was first popularised for image classification~\cite{hinton2015distilling} where a student model is trained to mimic the~\textit{soft labels} generated by a teacher model. However, this approach does not work well for object detection~\cite{wang2019fgfi} which consists of jointly classifying and localising objects. While soft label-based KD can be directly applied for classification, finding an equivalent for localisation remains a challenge. Recent work~\cite{dai2021gid,guo2021defeat,sun2020tadf,wang2019fgfi,yang2021fgd,zhang2021improve,zhixing2021frs} alleviates this problem by forcing the student model to generate feature maps similar to the teacher counterpart; a process known as \textit{feature imitation}. 

However, which features should the student imitate? This question is of the utmost importance for dense object detectors~\cite{chen2021ddod,li2020generalized,lin2017focal,tian2019fcos,zhang2020bridging,zhu2020autoassign} because, unlike two-stage detectors~\cite{cai2018cascade,he2017mask,ren2015faster}, they do not use the RoIAlign~\cite{he2017mask} operation to explicitly pool and align object features; instead they output predictions at every location of the feature map~\cite{li2019scale}.
Recent work~\cite{sun2020tadf,yang2021fgd} has shown that distilling the whole feature map with equal weighting is sub-optimal because not all features carry equally meaningful information. Therefore, a weighting mechanism that assigns appropriate importance to different regions, particularly to~\textit{foreground} regions near the objects, is highly desirable for dense object detectors, and has featured in recent work. For example, in DeFeat~\cite{guo2021defeat}, foreground features that lie within ground truth (GT) boxes (Fig.~\ref{fig:foreground}a) are distilled with equal weighting.  In~\cite{sun2020tadf} the authors postulate that useful features are located at the centre of \GT~boxes and weigh the foreground features using a Gaussian (Fig.~\ref{fig:foreground}b). In  Fine-grained Feature Imitation (FGFI)~\cite{wang2019fgfi}, the authors distil features covered by anchor boxes whose Intersection over Union (IoU) with the GTs are above a certain threshold (Fig.~\ref{fig:foreground}c).

\input{figures/tex/freground}
In this paper, we treat feature imitation for foreground regions differently. Instead of assigning distillation weights using hand-design policies, we argue that feature imitation should be conducted on a few \disregionIntro: the locations where the teacher model generates the most accurate predictions. Our intuition is that these regions should be distilled because they hold the information that leads to the best predictions; other areas will be less informative and can contaminate the distillation process by distracting from more essential features. To achieve our goal, we adapt the~\emph{quality} measure from~\cite{chen2021ddod} to score teacher predictions. Then, we conduct an experiment to visualise how these scores are distributed and verify that high-scoring~\disregion~contribute the most to teacher performance. Those findings drive us to propose a \ourmoduleIntro~module to weight the foreground distillation loss: inspired by recent progress in label assignment~\cite{zhang2020bridging,chen2021ddod,Ma2021iqdet,wang2021end,zhu2020autoassign} for dense detectors, we sample the top-K positions with the highest quality score from the teacher model and use an adaptive Gaussian distribution to fit the \disregion~for smoothly weighting the distillation loss. Fig.~\ref{fig:foreground}d shows a visual representation of the regions selected for distillation. We call our method \ourmethodIntro. Our contributions are as follows: 
\begin{enumerate}
    \item We conduct experiments to study how the~\emph{quality} scores of teacher predictions are distributed in the image plane and observe that the locations that make up the top-1\% of scores are responsible for most of the teacher's performance in modern state-of-the-art dense detectors. 
    \item Based on our observations, we propose using the \disregion~of the teacher as foreground features. We show that focusing distillation mainly on these few areas yields significant performance gains for the student model.
	\item We introduce a parameterless weighting scheme for foreground distillation pixels and show that when applied to our \disregion, we achieve even stronger distillation performance.
	\item We benchmark our approach on the COCO and CrowdHuman datasets and show its superiority over the state-of-the-art across multiple detectors.
\end{enumerate}

%% file: figures/tex/freground.tex
\begin{figure*}[!t]
    \centering
    \includegraphics[width=\textwidth]{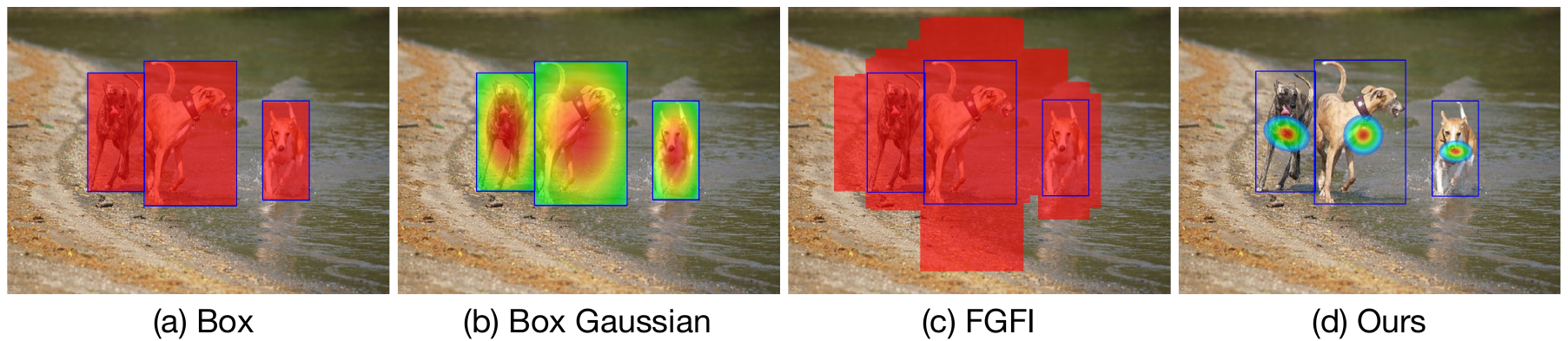}
    \captionsetup{font={footnotesize}}
    \caption{A comparison between different foreground distillation regions. The ground-truth bounding box is marked in blue. The colour heatmaps indicate the distillation weight for different areas. In contrast to other methods (a)-(c)~\cite{guo2021defeat,sun2020tadf,wang2019fgfi}, Our approach (d) focuses on a few key predictive regions of the teacher.}
    \label{fig:foreground}
\end{figure*}

%% file: sections/related_works.tex
\section{Related Work}

\noindent \textbf{Dense Object Detection.}
In the last few years, object detection has seen considerable gains in performance~\cite{cai2018cascade,carion2020end,chen2021ddod,he2017mask,li2020generalized,lin2017focal,RedmonDiGiFa16,redmon2017yolo9000,ren2015faster,tian2019fcos}. The demand for simple, fast models has brought one-stage detectors into the spotlight \cite{chen2021ddod,tian2019fcos}. In contrast to two-stage detectors, one-stage detectors directly regress and classify candidate bounding boxes from a pre-defined set of anchor boxes (or anchor points), alleviating the need for a separate region proposal mechanism. Anchor-based detectors \cite{chen2021ddod,lin2017focal} achieve good performance by regressing from anchor boxes with pre-defined sizes and ratios. In contrast, anchor-free methods~\cite{li2020generalized,tian2019fcos,zhu2020autoassign} regress directly from anchor points (or locations), eliminating the need for the additional hyper-parameters used in anchor-based models. A vital challenge for detectors is determining which bounding box predictions to label as positive and negative -- a problem frequently referred to as \emph{label assignment} \cite{zhu2020autoassign}. Anchors are commonly labelled as positives when their IoU with the \GT~is over a certain threshold (e.g. IoU $\geq$ 0.5) \cite{lin2017focal,tian2019fcos}, however, more elaborate mechanisms for label assignment have been proposed~\cite{tian2019fcos,zhang2020bridging,zhu2020autoassign,chen2021ddod}. For example, FCOS \cite{tian2019fcos} applies a weighting scheme to suppress low-quality positive predictions using a ``center-ness'' score. Other works dynamically adjust the number of positive instances according to statistical characteristics \cite{zhang2020bridging} or by using a differentiable confidence module \cite{zhu2020autoassign}. In DDOD~\cite{chen2021ddod}, the authors separate label assignment for the classification and regression branches and balance the influence of positive samples between different scales of the feature pyramid network (FPN).
\\

\noindent \textbf{Knowledge Distillation for Object Detection.}
Early KD approaches for classification focus on transferring knowledge to student models by forcing their predictions to match those of the teacher~\cite{hinton2015distilling}. More recent work~\cite{wang2019fgfi,yang2021fgd,zhixing2021frs} claims that feature imitation, i.e.\ forcing the intermediate feature maps of student models to match their teacher counterpart, is more effective for detection. A vital challenge when performing feature imitation for dense object detectors is determining which feature regions to distil from the teacher model. Naively distilling all feature maps equally results in poor performance \cite{guo2021defeat,sun2020tadf,yang2021fgd}. To solve this problem, FGFI~\cite{wang2019fgfi} distils features that are covered by anchor boxes which have a high IoU with the GT. However, distilling in this manner is still sub-optimal~\cite{dai2021gid,sun2020tadf,yang2021fgd,zheng2021ld,zhixing2021frs}. TADF~\cite{sun2020tadf} suppresses foreground pixels according to a static 2D Gaussian fitted over the \GT. LD~\cite{zheng2021ld} gives higher priority to central locations of the \GT~using DIoU~\cite{zheng2020distance}. GID~\cite{dai2021gid} propose to use the top-scoring predictions using L1 distance between the classifications scores of the teacher and the student, but do not account for location quality. In LAD~\cite{Nguyen2022lad}, the authors use~\textit{label assignment} distillation where the detector's encoded labels are used to train a student.  Others weight foreground pixels according to intricate adaptive weighting or attention mechanisms~\cite{dai2021gid,kang2021instance,yang2021fgd,yao2021gdetkd,zhixing2021frs}. However, these weighting schemes still heavily rely on the \GT~dimensions, and they are agnostic to the capabilities of the teacher. In contrast, we focus distillation on only a few~\disregion~using a combination of classification and regression scores as a measure of quality. We then smoothly aggregate and weigh the selected locations using an estimated 2D Gaussian, which further focuses distillation and improves performance. This allows us to dynamically adjusts to different sizes and orientations of objects independently of the \GT~dimensions while accounting for the teacher's predictive abilities. 
 

%% file: sections/method_v2.tex
\section{Method}
We begin by describing how to measure the predictive quality of a bounding box prediction and find the~\disregionIntro~of a teacher network (Sec.~\ref{sec:pred_regions}). Then, we introduce our \ourmoduleIntro~module that returns a foreground distillation mask based on these regions~(Sec.~\ref{sec:distill_regions}). Finally, we describe our full Prediction-Guided Distillation pipeline (Sec.~\ref{sec:distill_pipeline}).

\input{figures/tex/criteria}

\subsection{Key Predictive Regions}\label{sec:pred_regions}
Our goal is to amplify the distillation signal for the most meaningful features produced by a teacher network. For this purpose, we look at the~\emph{quality} of a teacher's bounding box predictions taking both classification and localisation into consideration, as defined in~\cite{chen2021ddod}. Formally, the quality score of a box $\hat{b}_{(i,j)}$ predicted from a position $X_i=(x_i, y_i)$ w.r.t. a ground truth $b$ is:
\begin{equation}
\begin{footnotesize}
	\begin{aligned}
        q(\hat{b}_{(i,j)}, b) = \underbrace{\mathds{1} \left[ X_i \in b \right]}_{\text{indicator}} ~\cdot~ \underbrace{{\Bigl( \hat{p}_{(i,j)}(b) \Bigr)}^{1-\xi}}_{\text{classification}} ~\cdot \underbrace{{\Bigl( \mathrm{IoU} \bigl( b, \hat{b}_{(i,j)} \bigr) \Bigr)}^\xi}_{\text{localisation}}
    \label{eq:quality}
    \end{aligned}
\end{footnotesize}
\end{equation} where $\mathds{1} \left[ X_i \in \Omega_b \right]$ is an indicator function that is 1 if $X_i$ lies inside box $b$ and 0 otherwise; $\hat{p}_{(i,j)}(b)$ is the classification probability w.r.t. the \GT~box's category; $\mathrm{IoU} \bigl( b, \hat{b}_{(i,j)} \bigr)$ is the IoU between the predicted and ground-truth box; $\xi$ is a hyper-parameter that balances classification and localisation. We calculate the quality score of location $X_i$ as the maximum value of all prediction scores for that particular location, i.e. $\hat{q}_{i} = max_{j\in J_i}~{q(\hat{b}_{(i,j)}, b)}$, where $J_i$ is the set of predictions at location $X_i$. While this quality score has been applied for standard object detection~\cite{chen2021ddod}, we are the first to use it to identify useful regions for distillation.

In Fig.~\ref{fig:criteria} we visualise the heatmaps of prediction quality scores for five state-of-the-art detectors, including anchor-based (ATSS~\cite{zhang2020bridging} and DDOD~\cite{chen2021ddod}) and anchor-free (FCOS~\cite{tian2019fcos}, GFL~\cite{li2020generalized} and AutoAssign~\cite{zhu2020autoassign}) detectors. Across all detectors, we observe some common characteristics: 
(1) For the vast majority of objects, high scores are concentrated around a {\bf single region}; 
(2) The size of this region doesn't necessarily correlate strongly with the size of the actual \GT~box;
(3) Whether or not the centring prior~\cite{tian2019fcos,zhu2020autoassign} is applied for label assignment during training, this region tends to be close to the centre of the \GT~box. 
These observations drive us to develop a~\ourmoduleIntro~ module to focus the distillation on these important regions.

\subsection{Prediction-Guided Weighting Module}\label{sec:distill_regions}
The purpose of KD is to allow a student to mimic a teacher's strong generalisation ability. To better achieve this goal, we propose to focus foreground distillation on locations where a teacher model can yield predictions with the highest quality scores because those locations contain the most valuable information for detection and are critical to a teacher's high performance. In Fig.~\ref{fig:quality_full} we present the results of a pilot experiment to identify how vital these high-scoring locations are for a detector. Specifically, we measure the performance of different pre-trained detectors after masking out their top-$X\%$ predictions before non-maximum suppression (NMS) during inference. We observe that in all cases the mean Averaged Precision (mAP) drops dramatically as the mask-out ratio increases. Masking out the top-1\% of predictions incurs around a 50\% drop in AP. This suggests that the \disregionIntro~(responsible for the majority of a dense detector's performance) lie within the top-1\% of all anchor positions bounded by the \GT~box.

\input{figures/tex/quality_full}

Given their significance, how do we incorporate these regions into distillation? We could simply use all feature locations weighted by their quality score, however, as we show in Sec.~\ref{sec:ablation} this does not yield the best performance. Inspired by recent advances in label assignment for dense object detectors~\cite{chen2021ddod,wang2021end}, we instead propose to focus foreground distillation on the top-K positions (feature pixels) with the highest quality scores across all FPN levels. We then smooth the influence of each position according to a 2D Gaussian distribution fitted by Maximum-Likelihood Estimation (MLE) for each \GT~box. Finally, foreground distillation is conducted only on those K positions with their weights assigned by the Gaussian.

Formally, for an object $o$ with \GT~box $b$, we first compute the quality score for each feature pixel inside $b$, then we select the K pixels with the highest quality score $T^{o}=\{(X^o_k, l^o_k)|k=1,...,K\}$ across all FPN levels, in which $X^o_k$ and $l^o_k$ are the absolute coordinate and the FPN level of the $k$-th pixel. Based on our observation in Sec.~\ref{sec:pred_regions}, we assume the selected pixels $T^o_{k}$ are drawn as $T^o_k\sim \mathcal{N}(\mu,\Sigma|o)$ defined on the image plane and use MLE to estimate $\mu$ and $\Sigma$:
\begin{footnotesize}
\begin{align}
    \hat{\mu} = \frac{1}{K}\sum_{k=1}^{K}X^o_k, ~~\hat{\Sigma} = \frac{1}{K}\sum_{k=1}^{K}(X^o_k-\hat{\mu})(X^o_k-\hat{\mu})^T
\end{align}
\end{footnotesize}Then, for every feature pixel $P_{(i,j),l}$ on FPN layer $l$ with absolute coordinate $X_{i,j}$, we compute its distillation importance w.r.t. object $o$ by:
\begin{equation}
\begin{footnotesize}
\begin{aligned}
    I^o_{(i,j),l} = \begin{cases} 
                 ~~~~~~~~~~~~~~~~~~~~~~0 & P_{(i,j),l} \notin T^{o} \\
                 \exp{\bigl( -\frac{1}{2}(X_{i,j}-\hat{\mu})\hat{\Sigma}^{-1}(X_{i,j}-\hat{\mu})^T \bigr)} & P_{(i,j),l} \in T^{o} 
                 \end{cases}
\end{aligned}
\end{footnotesize}
\end{equation}
If a feature pixel has non-zero importance for multiple objects, we use its maximum: $I_{(i,j),l}=\max_o{\{I^o_{(i,j),l}\}}$. Finally, for each FPN level $l$ with size $H_l \times W_l$, we assign the distillation weight $M_{(i,j),l}$ by normalising the distillation importance by the number of non-zero importance pixels at that level:
\begin{footnotesize}
\begin{align}\label{eq:foreweight}
    \mathbf{M_{(i,j),l}} = \frac{I_{(i,j),l}}{\sum_{i=1}^{H_l}\sum_{j=1}^{W_l}\mathds{1}_{(i,j),l}}
\end{align}
\end{footnotesize}where $\mathds{1}_{(i,j),l}$ is an indicator function that outputs 1 if $I_{(i,j),l}$ is not zero. The process above constitutes our~\ourmoduleIntro~module whose output is a foreground distillation weight $\mathbf{M}$ across all feature levels and pixels. 

\input{figures/tex/pipeline}

\subsection{Prediction-Guided Distillation}
\label{sec:distill_pipeline}
In this section, we introduce our KD pipeline, which is applicable to any dense object detector. We build our work on top of the state-of-the-art Focal and Global Distillation (FGD)~\cite{yang2021fgd} and incorporate their spatial and channel-wise attention mechanisms. In contrast to other distillation methods, we use the output mask from our \ourmodule~module to focus the distillation loss on the most important foreground regions. Moreover, we decouple the distillation for the classification and regression heads to better suit the two different tasks \cite{oksuz2020ranking,chen2021ddod}. An illustration of the pipeline is shown in Fig~\ref{fig:pipeline}.
\\

\noindent \textbf{Distillation of Features.}
We perform feature imitation at each FPN level, encouraging feature imitation on the first feature maps of the regression and classifications heads. Taking inspiration from \cite{chen2021ddod}, we separate the distillation process for the classification and regression heads -- distilling features of each head independently. Formally, at each feature level of the FPN, we generate two foreground distillation masks $\mathbf{M^{cls}},\mathbf{M^{reg}}\in \mathds{R}^{H \times W}$ with different $\xi^{cls}$ and $\xi^{reg}$ using \ourmodule. Then, student features $F^{S, cls}, F^{S, reg}\in \mathds{R}^{C \times H \times W}$ are encouraged to mimic teacher features $F^{T, cls}, F^{T, reg}\in \mathds{R}^{C \times H \times W}$ as follows:

\begin{footnotesize}
\begin{align}\label{eql:distill}
&L_{fea}^{cls} =  \sum_{k=1}^{C}\sum_{i=1}^{H}\sum_{j=1}^{W}  (\alpha \mathbf{M_{i,j}^{cls}}+\beta  N_{i,j}^{cls})P_{i,j}^{T,cls}A_{k,i,j}^{T,cls}(F^{T,cls}_{k,i,j}-F^{S,cls}_{k,i,j})^2 \\
&L_{fea}^{reg} =   \sum_{k=1}^{C}\sum_{i=1}^{H}\sum_{j=1}^{W}  \gamma \mathbf{M_{i,j}^{reg}}A_{k}^{T,reg}(F^{T,reg}_{k,i,j}-F^{S,reg}_{k,i,j})^2 
\end{align}
\end{footnotesize}where $\alpha,\beta,\gamma$ are hyperparameters to balance between loss weights; $N^{cls}$ is the normalised mask over background distillation regions: $N^{cls}_{i,j}={\mathds{1}^{-}_{i,j}}/{\sum_{h=1,w=1}^{H,W}\mathds{1}^{-}_{w,h}}$ where $\mathds{1}^{-}_{a,b}$ is the background indicator that becomes 1 if pixel $(a,b)$ does not lie within any \GT~box. $P$ and $A$ are spatial and channel attention maps from \cite{yang2021fgd} as defined below. Note, we do not use the Global Distillation Module in FGD and the adaptation layer that is commonly used in many KD methods~\cite{zhixing2021frs,wang2019fgfi,zhang2021improve,chen2017learning,guo2021defeat,yang2021fgd} as we find them have negligible impact to the overall performance. 
\\

\noindent \textbf{Distillation of Attention.}
We build on the work in FGD \cite{yang2021fgd} and additionally encourage the student to imitate the attention maps of the teacher. We use spatial attention as defined in \cite{yang2021fgd}, but we modify their channel attention by computing it independently for each feature location instead of all spatial locations. Specifically, we define s\textbf{p}atial attention $\mathbf{P} \in \mathds{R}^{1 \times H \times W}$ and ch\textbf{a}nnel attention $\mathbf{A} \in \mathds{R}^{C \times H \times W}$ over a single feature map $F \in \mathds{R}^{C \times H \times W}$ as follows:
\begin{footnotesize}
\begin{align}
P_{i,j} =  \frac{HW \cdot \exp{(\sum_{k=1}^{C}|F_{k,i,j}}|/\tau)}{\sum_{i=1}^{H}\sum_{j=1}^{W}\exp{(\sum_{k=1}^{C}|F_{k,i,j}|/\tau)}},~~~~
A_{k,i,j} = \frac{C \cdot \exp{(|F_{k,i,j}|/\tau)}}{\sum_{k=1}^{C}\exp{(|F_{k,i,j}|/\tau)}}
\end{align} 
\end{footnotesize}Similar to feature distillation, we decouple the attention masks for classification and regression for the teacher and student: $A^{T,cls}$, $A^{T,reg}$, $P^{S,cls}$. The two attention losses are defined as follows: 
\begin{footnotesize}
\begin{align}
    &L_{att}^{cls} = \frac{\delta}{HW}\sum_{i=1}^{H}\sum_{j=1}^{W}|P_{i,j}^{T,cls}-P_{i,j}^{S,cls}| + \frac{\delta}{CHW}\sum_{k=1}^{C}\sum_{i=1}^{H}\sum_{j=1}^{W}|A_{k,i,j}^{T,cls}-A_{k,i,j}^{S,cls}| \\
    &L_{att}^{reg} = \frac{\delta}{C\sum_{i=1}^{H}\sum_{j=1}^{W}\mathds{1}_{i,j}}\sum_{i=1}^{H}\sum_{j=1}^{W}\sum_{k=1}^{C}\mathds{1}_{i,j}|A_{k,i,j}^{T,reg}-A_{k,i,j}^{S,reg}| 
\end{align}
\end{footnotesize}where $\delta$ is balancing loss weight hyperparameter; and $\mathds{1}_{i,j}$ is an indicator that becomes 1 when $\mathbf{M^{reg}_{i,j}} \neq 0$.
\\

\noindent \textbf{Full Distillation.} The full distillation loss is 
\begin{footnotesize}
\begin{align}\label{eql:distillAll}
L_{distill} = L_{fea}^{cls} + L_{fea}^{reg} + L_{att}^{cls} + L_{att}^{reg}
\end{align}
\end{footnotesize}

%% file: figures/tex/criteria.tex
\begin{figure*}[!t]
    \centering
    \includegraphics[width=\textwidth]{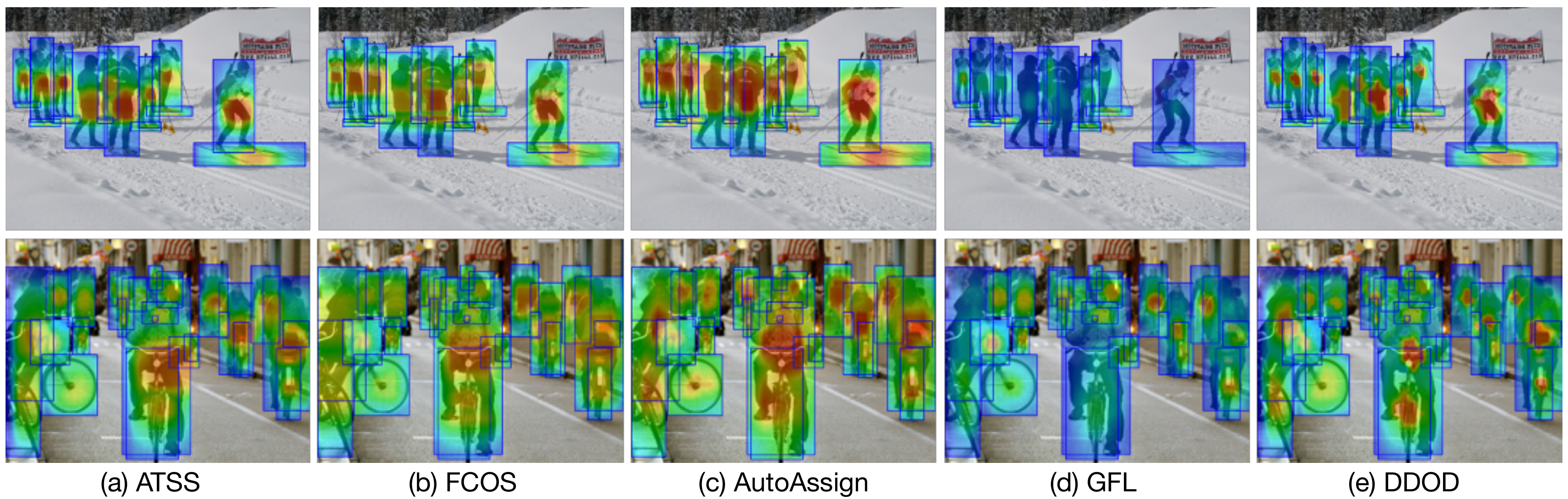}
    \captionsetup{font={footnotesize}}
    \caption{A visualisation of quality scores for various dense object detectors with $\xi=0.8$ following~\cite{chen2021ddod}. We acquire the quality heatmap by taking the maximum value at each position across FPN layers.}
    \label{fig:criteria}
\end{figure*}

%% file: figures/tex/quality_full.tex
\begin{SCfigure*}[][!t]
  \includegraphics[width=0.5\textwidth]{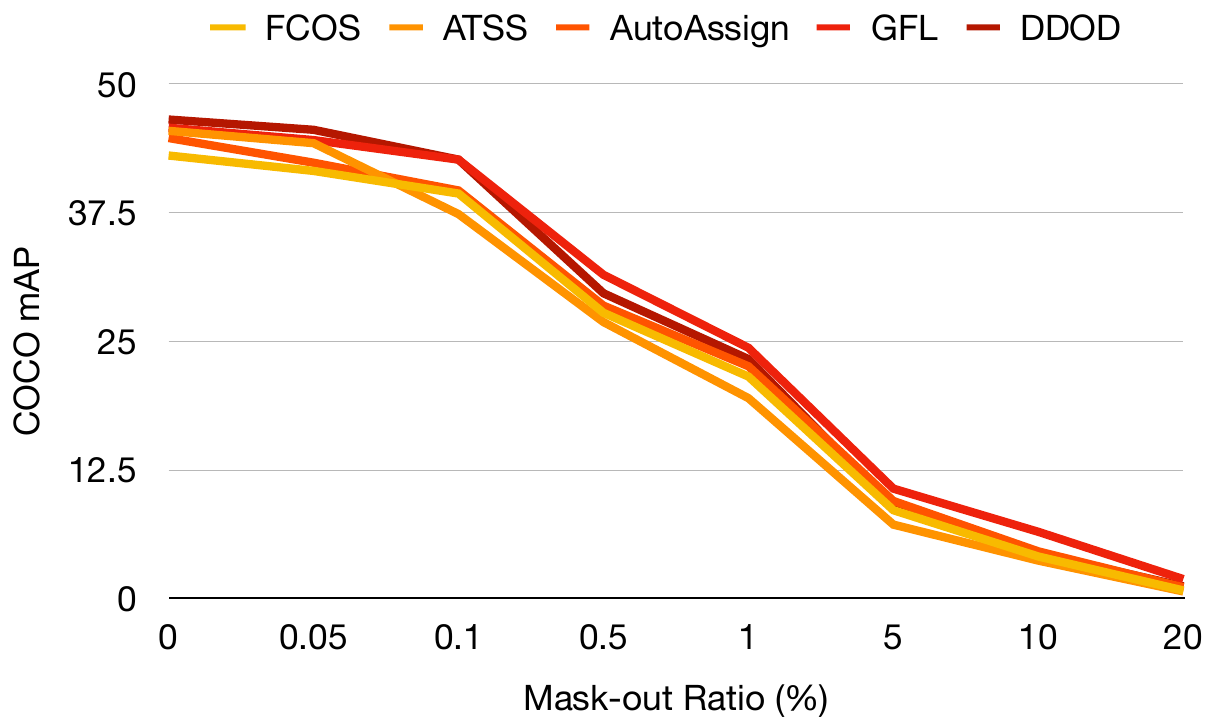}
  \captionsetup{font={footnotesize}}
  \caption{COCO mAP performance of pre-trained detectors after ignoring predictions in the top-X\% of quality scores during inference. We observe that the top-1\% predictions within the \GT~box region are responsible for most performance gains.}
  \label{fig:quality_full}
\end{SCfigure*}

%% file: figures/tex/pipeline.tex
\begin{figure*}[!t]
    \centering
    \includegraphics[width=\textwidth]{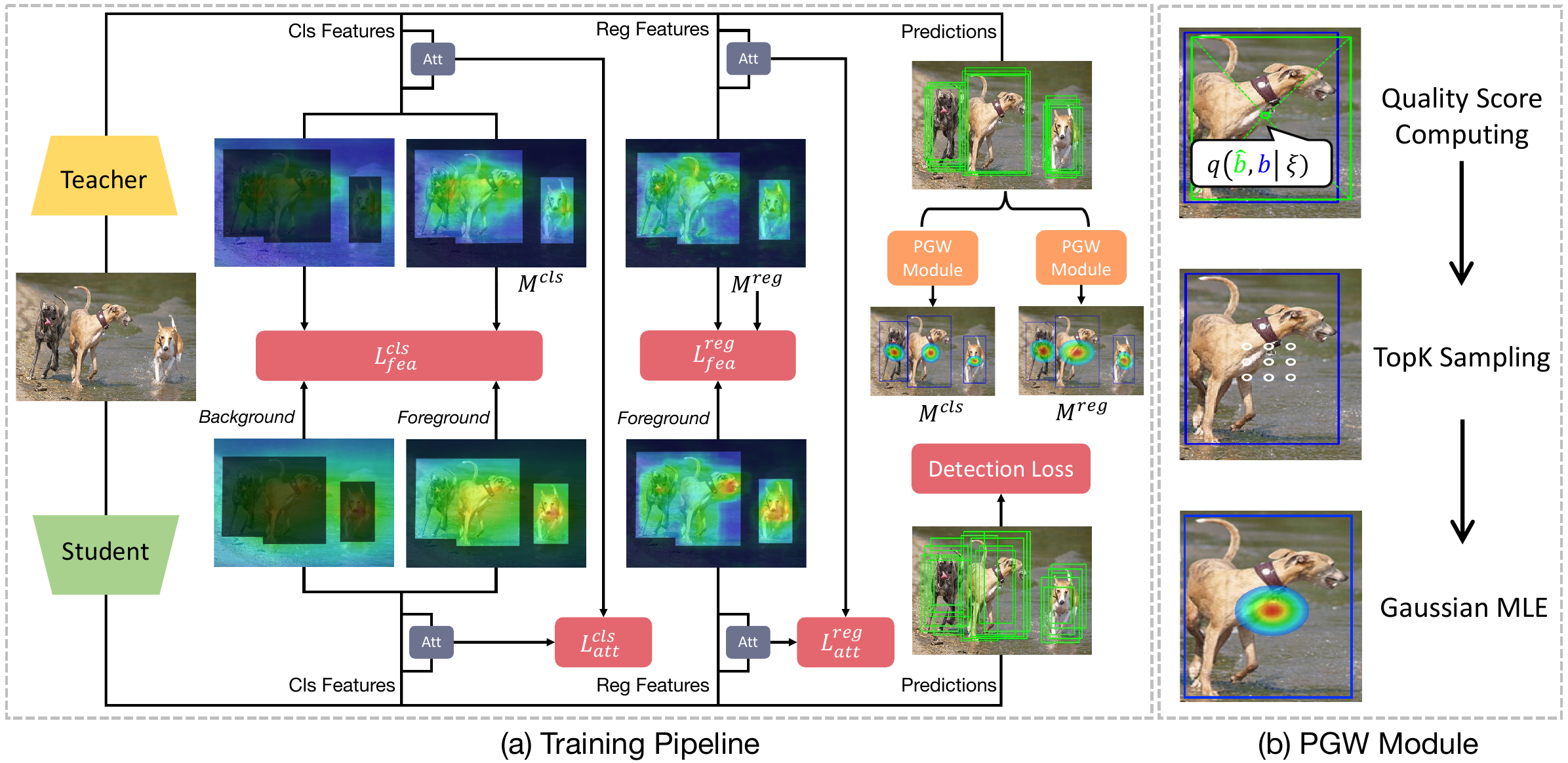}
    \captionsetup{font={footnotesize}}
\caption{Our~\ourmethodIntro~pipeline. The~\ourmoduleIntro~modules find the teacher's~\disregion~and generates a foreground distillation weighting mask by fitting a Gaussian over these regions. Our pipeline also adopts the attention masks from FGD~\cite{yang2021fgd} and distils them together with the features. We distil the classification and regression heads separately to accommodate for these two distinct tasks. \cite{chen2021ddod}.}
\label{fig:pipeline}
\end{figure*}

%% file: sections/experiments.tex
\section{Experiments}

\input{tabels/big_compare}

\subsection{Setup and Implementation Details}\label{sec:implementation}
We evaluate \ourmethod~on two benchmarks: COCO~\cite{lin2014microsoft} for general object detection and CrowdHuman~\cite{shao2018crowdhuman} for crowd scene detection; this contains a large number of heavily occluded objects. Our codebase is built on PyTorch~\cite{paszke2019pytorch} and the MMDetection~\cite{chen2019mmdetection} toolkit and is available at~\url{https://github.com/ChenhongyiYang/PGD}. All models are trained on 8 Nvidia 2080Ti GPUs. For both COCO and CrowdHuman, all models are trained using batch sizes of 32 and with an initial learning rate of 0.02, we adopt ImageNet pre-trained backbones and freeze all Batch Normalisation layers during training. Unless otherwise specified, on both dataset we train teacher models for 3$\times$ schedule (36 epochs)~\cite{he2019rethinking} with multi-scale inputs using ResNet-101~\cite{he2016deep} as backbone, and train student models for 1$\times$ schedule (12 epochs) with single-scale inputs using ResNet-50 as backbone. The COCO models are trained using the \textit{train2017} set and evaluated on \textit{mini-val} set following the official evaluation protocol~\cite{lin2014microsoft}. The CrowdHuman models are trained using the CrowdHuman \textit{training} set, which are then evaluated on the CrowdHuman \textit{validation} set following~\cite{Chu2020crowddet}. We set K in the top-K operation to 30 for all detectors and set $\alpha$ to 0.8 and 0.4 for anchor-based and anchor-free detectors respectively. Following~\cite{yang2021fgd}, we set $\sigma=0.0008$, $\tau=0.8$ and $\beta=0.5\alpha$; we set $\xi^{cls}=0.8$ and $\xi^{reg}=0.6$ following \cite{chen2021ddod}. We empirically set $\gamma=1.6\alpha$ with minimal tuning.

\input{tabels/small_backbone}

\subsection{Main Results}
\noindent \textbf{Comparison with State-of-the-art.} We compare our \ourmethod~and other recent state-of-the-art object detection KD approaches for five high-performance dense detectors on COCO; these are a mixture of anchor-based (ATSS and DDOD) and anchor-free (FCOS, GFL and AutoAssign) detectors for COCO. The results are presented in Table~\ref{tbl:big-compare}. We use the same teacher and student models and the same training settings in each case, and all training is conducted locally. For competing distillation methods, we follow the hyper-parameter settings in their corresponding papers or open-sourced code repositories. We observe that our methods surpass other KD methods with a large margin for all five detectors, which validates the effectiveness of our approach. Our approach significantly improvement over the baseline approach FGD~\cite{yang2021fgd} and even outperforms LD~\cite{zheng2021ld} when applied to GFL~\cite{li2020generalized}, which was specifically designed for this detector. We observe \ourmethod~is particularly good at improving the AP$_{75}$ of student models, suggesting that the student model's localisation abilities have been largely improved.
\\

\input{tabels/crowdhuman}

\noindent \textbf{Distilling to a Lightweight Backbone.} Knowledge Distillation is usually used to transfer useful information from a large model to a lightweight model suitable for deployment on the edge. With this in mind, we apply \ourmethod~using a ResNet-101 as the teacher backbone and a MobileNet V2~\cite{sandler2018mobilenetv2} as the student backbone on anchor-based (ATSS) and anchor-free (FCOS) detectors. The results are provided in Tab.~\ref{tbl:lightweighted}. Our method surpasses the baseline by a significant margin, pointing to its potential for resource-limited applications.
\\

\noindent \textbf{Distillation for Crowd Detection.} We compare our approach to other KD methods on the challenging CrowdHuman dataset that features heavily crowded scenes. We use the DDOD object detector for this experiment as it achieves the strongest performance. In addition to detection AP, we report the log miss rate (MR)~\cite{Chu2020crowddet} designed for evaluation in crowded scenes as well as the Jaccard Index (JI) that evaluates a detector's counting ability. The results are available in Table~\ref{tbl:crowdhuman}. Our approach performs better than all competing methods. While FGD achieves comparable MR and JI scores to our method, the AP for our methods is significantly greater. We believe this is because \ourmethod~strongly favours highly accurate predictions during distillation, which directly impacts the AP metric. 
\\

\input{tabels/selfdistill}

\noindent \textbf{Self-Distillation.} Self-distillation is a special case of knowledge distillation where the teacher and student models are exactly same. It is useful as it can boost a model's performance while avoiding introducing extra parameters. We compare the our method's self-distillation performance with the baseline FGD and present results for both anchor-free FCOS and anchor-based ATSS in Tab.~\ref{tbl:selfdistill}. The teachers and students use ResNet-50 as backbone and are trained with 1$\times$ schedule using single-scale inputs. We can see that our approach achieves a better performance than the baseline, indicating its effectiveness in self-distillation.

\subsection{Ablation Study}\label{sec:ablation}
\input{tabels/foreground}
\input{tabels/hyperparameter}
\noindent \textbf{Comparing Foreground Distillation Strategies.} We compare alternative strategies for distilling foreground regions to investigate how important is distilling different foreground regions. We use ATSS as our object detector and present results in Tab.~\ref{tbl:foreground}. Note here we only modify the foreground distillation strategy while keeping everything else the same. We first evaluate the strategy used in FGD~\cite{yang2021fgd} and DeFeat~\cite{guo2021defeat}, where regions in the \GT~box are distilled equally. We dub this the~\textit{Box} strategy (Fig.~\ref{fig:foreground}a). Compared to our method, \textit{Box} achieves 0.9 AP worse performance. A possible reason for this is that it can include sub-optimal prediction locations that distract from more meaningful features. Note that the \textit{Box} strategy still outperforms the baseline FGD, we attribute this improvement to the decoupling of distillation for classification and regression branches. Several works \cite{zheng2021ld,sun2020tadf} postulate that most meaningful regions lie near the centre of the \GT~box. We evaluate the \textit{BoxGauss} strategy that was proposed in TADF~\cite{sun2020tadf} (Fig.~\ref{eq:foreweight}b). Specifically, a Gaussian distribution is used to weight the distillation loss, where its mean is the centre of the \GT~box, and the standard deviation is calculated from the box dimensions. This strategy yields +0.4 AP improvement over vanilla \textit{Box} strategy, suggesting the importance of focusing on the centre area; however, it is still surpassed by our approach. We consider a \textit{Centre} strategy, which distils a $ 0.2 H \times 0.2 W$ area at the middle of the GT box. Somewhat surprisingly, this achieves an even worse AP than the vanilla \textit{Box} strategy in almost all instances, with comparable performance on small objects. A possible explanation is that a fixed ratio region fails to cover the full span of useful regions for different-sized objects and limits the amount of distilled information. Then we compare to an adaptive loss weighting mechanism where we directly use the quality score in Equation~\ref{eq:quality} to weight features for the distillation loss. The strategy---which we refer to as~\textit{Quality}---improves slightly on \textit{BoxGauss}, especially for medium and high scoring boxes. However, it significantly under-performs on small objects.
In contrast, the \textit{TopkEq} strategy, where we limit distillation to only the top-K pixels according to the quality score (we set $K=30$ to match our method), provides a significant improvement to the detection of small objects. A possible explanation for this is that distilling on positions with lower scores still introduces considerable noise, whereas limiting distillation to only the highest-scoring pixels focuses the student towards only the most essential features of the teacher. Finally, we compare our method to one that replaces the Gaussian MLE with kernel density estimation, the \textit{KDE} strategy. It achieves similar performance to our Gaussian MLE approach, but is more complicated.
\\

\noindent \textbf{Hyper-parameter Settings.} 
Here, we examine the effect of changing two important hyper-parameters used in our approach, as presented in Tab.~\ref{tbl:hyperparameter}. The first is $K$, which is the number of high-scoring pixels used for distillation. The best performance is obtained for $K=30$. Small $K$ can cause distillation to neglect important regions, whereas large $K$ can introduce noise that distracts the distillation process from the most essential features. The second hyper-parameter we vary is $\alpha$ which controls the magnitude of the distillation loss. We can see how this affects performance for anchor-based ATSS and anchor-free FCOS. We find that the ATSS's performance is quite robust when $\alpha$ is between 0.05--0.1, and FCOS can achieve good performance when $\alpha$ is between 0.03--0.1. For both types of detectors, a small $\alpha$ will minimise the effect of distillation, and a large $\alpha$ can make training unstable.
\\

\noindent \textbf{Decoupled Distillation.} In our pipeline, we decouple the KD loss to distil the classification and regression heads separately (see Section~\ref{sec:distill_pipeline}). This practice differs from previous feature imitation-based approaches where the FPN neck features are distilled. Here we conduct experiments to test this design and present the result in Tab.~\ref{tbl:branch}. Firstly, we remove the regression KD loss and only apply the classification KD loss using FPN features. The model achieves 43.6 mAP on COCO. Then we test only applying the classification KD loss using the classification feature map; the performance improves very slightly (by 0.2). Next, we only test the regression KD loss using the regression features, resulting in 41.7 COCO mAP. The performance is significantly harmed because the regression KD loss only considers foreground regions while ignoring background areas. Finally, we come to our design by combining both classification and regression KD losses, which achieves the best performance, at 44.2 COCO mAP.
\\

\input{figures/tex/qulitative}
\input{tabels/distill_branch}

\noindent \textbf{Qualitative Studies.}
We visualise box predictions using ATSS as our object detector in Fig.~\ref{fig:qualitative}, in which we show \GT~boxes alongside student predictions with and without distillation using \ourmethod. While the high-performance ATSS is able to accurately detect objects in most cases, we observe some clear advantages of using our distillation approach: it outputs fewer false positives (Fig.~\ref{fig:qualitative} b,c), improves detection recall (Fig.~\ref{fig:qualitative} a,d), and localises objects better (Fig.~\ref{fig:qualitative} b,d,e).

%% file: tabels/big_compare.tex
\begin{table*}[ht!]
\renewcommand\arraystretch{1.2}
\scriptsize \setlength{\tabcolsep}{4.5pt}
  \centering
  \begin{threeparttable}
    \begin{tabularx}{\textwidth}{X|X|XXXXXX}
        \hline   
        \rowcolor{white}
        Detector & Setting & AP & AP$_{50}$ & AP$_{75}$ & AP$_{S}$ & AP$_{M}$ & AP$_{L}$ \\
        \hline 
        \multirow{7}{*}{FCOS\cite{tian2019fcos}} & \cellcolor{gray!25}Teacher  &\cellcolor{gray!25}43.1 &\cellcolor{gray!25}62.4 &\cellcolor{gray!25}46.6 &\cellcolor{gray!25}25.5 &\cellcolor{gray!25}47.1 &\cellcolor{gray!25}54.7 \\
                            & \cellcolor{gray!25}Student  &\cellcolor{gray!25}38.2 &\cellcolor{gray!25}57.9 &\cellcolor{gray!25}40.5 &\cellcolor{gray!25}23.1 &\cellcolor{gray!25}41.3 &\cellcolor{gray!25}49.4 \\
                            & DeFeat~\cite{guo2021defeat}  &40.7(+2.5) &60.5(+2.6) &43.5(+3.0) &24.7(+1.6) &44.4(+3.1) &52.4(+3.0) \\
                            & FRS\cite{zhixing2021frs} &40.9(+2.7) &60.6(+2.7) &44.0(+3.5) &25.0(+1.9) &44.4(+3.1) &52.6(+3.2) \\
                            & FKD\cite{zhang2021improve}     &41.3(+3.1) &60.9(+3.0) &44.1(+3.6) &23.9(+0.8) &44.9(+3.6) &53.8(+4.4) \\
                            & FGD\cite{yang2021fgd}     &41.4(+3.2) &61.1(+3.2) &44.2(+3.7) &25.3(+\textbf{2.2}) &45.1(+3.8) &53.8(+4.4) \\
                            & Ours    &42.5(+\textbf{4.3}) &62.0(+\textbf{4.1}) &45.4(+\textbf{4.9}) &24.8(+1.7) &46.1(+\textbf{5.8}) &55.5(+\textbf{6.1}) \\
        \hline
        \multirow{7}{*}{Auto-} & \cellcolor{gray!25}Teacher  &\cellcolor{gray!25}44.8 &\cellcolor{gray!25}64.1 &\cellcolor{gray!25}48.9 &\cellcolor{gray!25}27.3 &\cellcolor{gray!25}48.8 &\cellcolor{gray!25}57.5 \\
        \multirow{7}{*}{Assign\cite{zhu2020autoassign}} & \cellcolor{gray!25}Student  &\cellcolor{gray!25}40.6 &\cellcolor{gray!25}60.1 &\cellcolor{gray!25}43.8 &\cellcolor{gray!25}23.6 &\cellcolor{gray!25}44.3 &\cellcolor{gray!25}52.4 \\
                            & DeFeat~\cite{guo2021defeat}  &42.3(+1.7) &61.6(+1.5) &46.1(+2.3) &24.1(+0.5) &46.0(+1.7) &54.4(+2.0) \\
                            & FRS~\cite{zhixing2021frs}     &42.4(+1.8) &61.9(+1.8) &46.0(+2.2) &24.9(+1.3) &46.0(+1.7) &54.8(+2.4) \\
                            & FKD\cite{zhang2021improve}     &42.8(+2.2) &62.1(+2.0) &46.5(+2.7) &25.7(+2.1) &46.4(+2.1) &55.5(+3.1) \\
                            & FGD\cite{yang2021fgd}     &43.2(+2.6) &62.5(+2.4) &46.9(+3.1) &25.2(+1.6) &46.7(+2.4) &56.2(+3.8) \\
                            & Ours    &43.8(+\textbf{3.1}) &62.9(+\textbf{2.8}) &47.4(+\textbf{3.6}) &25.8(+\textbf{2.2}) &47.3(+\textbf{3.0}) &57.5(+\textbf{5.1}) \\
        \hline
        \multirow{8}{*}{ATSS\cite{zhang2020bridging}} & \cellcolor{gray!25}Teacher  &\cellcolor{gray!25}45.5 &\cellcolor{gray!25}63.9 &\cellcolor{gray!25}49.7 &\cellcolor{gray!25}28.7 &\cellcolor{gray!25}50.1 &\cellcolor{gray!25}57.8 \\
                            & \cellcolor{gray!25}Student  &\cellcolor{gray!25}39.6 &\cellcolor{gray!25}57.6 &\cellcolor{gray!25}43.2 &\cellcolor{gray!25}23.0 &\cellcolor{gray!25}42.9 &\cellcolor{gray!25}51.2 \\
                            & DeFeat~\cite{guo2021defeat}  &41.8(+2.2) &60.3(+2.7) &45.3(+2.1) &24.8(+1.8) &45.6(+2.7) &53.5(+2.3) \\
                            & FRS~\cite{zhixing2021frs}     &41.6(+2.0) &60.1(+2.5) &44.8(+1.6) &24.9(+1.9) &45.2(+2.3) &53.2(+2.0) \\
                            & FGFI\cite{wang2019fgfi}    &41.8(+2.2) &60.3(+2.7) &45.3(+2.1) &24.8(+1.8) &45.6(+2.7) &53.5(+2.3) \\
                            & FKD\cite{zhang2021improve}     &42.3(+2.7) &60.7(+3.1) &46.2(+3.0) &26.3(+3.3) &46.0(+3.1) &54.6(+3.4) \\
                            & FGD\cite{yang2021fgd}     &42.6(+3.0) &60.9(+3.3) &46.2(+3.0) &25.7(+2.7) &46.7(+3.8) &54.5(+3.3) \\
                            & Ours    &44.2(+\textbf{4.6}) &62.3(+\textbf{4.7}) &48.3(+\textbf{5.1}) &26.5(+\textbf{3.5}) &48.6(+\textbf{5.7}) &57.1(+\textbf{5.9}) \\
        \hline
        \multirow{8}{*}{GFL\cite{li2020generalized}} & \cellcolor{gray!25}Teacher  &\cellcolor{gray!25}45.8 &\cellcolor{gray!25}64.2 &\cellcolor{gray!25}49.8 &\cellcolor{gray!25}28.3 &\cellcolor{gray!25}50.3 &\cellcolor{gray!25}58.6 \\
                            & \cellcolor{gray!25}Student  &\cellcolor{gray!25}40.2 &\cellcolor{gray!25}58.4 &\cellcolor{gray!25}43.3 &\cellcolor{gray!25}22.7 &\cellcolor{gray!25}43.6 &\cellcolor{gray!25}52.0 \\
                            & DeFeat~\cite{guo2021defeat}  &42.1(+1.9) &60.5(+2.1) &45.2(+1.9) &24.4(+1.7) &46.1(+2.5) &54.5(+2.5) \\
                            & FRS~\cite{zhixing2021frs}     &42.2(+2.0) &60.6(+2.2) &45.6(+2.3) &24.7(+2.0) &46.0(+2.4) &55.5(+3.5) \\
                            & FKD\cite{zhang2021improve}     &43.1(+2.9) &61.6(+3.2) &46.6(+3.3) &25.1(+2.4) &47.2(+3.6) &56.5(+4.5) \\
                            & FGD\cite{yang2021fgd}     &43.2(+3.0) &61.8(+3.4) &46.9(+3.6) &25.2(+2.5) &47.5(+3.9) &56.2(+4.2) \\
                            & LD\cite{zheng2021ld}      &43.5(+3.3) &61.8(+3.4) &47.4(+\textbf{4.1}) &24.7(+2.0) &47.5(+3.9) &57.3(+5.3) \\
                            & Ours    &43.8(+\textbf{3.6}) &62.0(+\textbf{3.6}) &47.4(+\textbf{4.1}) &25.4(\textbf{+2.7}) &47.8(+\textbf{4.2}) &57.6(+\textbf{5.6}) \\
        \hline
        \multirow{8}{*}{DDOD\cite{chen2021ddod}} & \cellcolor{gray!25}Teacher  &\cellcolor{gray!25}46.6 &\cellcolor{gray!25}65.0 &\cellcolor{gray!25}50.7 &\cellcolor{gray!25}29.0 &\cellcolor{gray!25}50.5 &\cellcolor{gray!25}60.1 \\
                            & \cellcolor{gray!25}Student  &\cellcolor{gray!25}42.0 &\cellcolor{gray!25}60.2 &\cellcolor{gray!25}45.5 &\cellcolor{gray!25}25.7 &\cellcolor{gray!25}45.6 &\cellcolor{gray!25}54.9 \\
                            & DeFeat~\cite{guo2021defeat}  &43.2(+1.2) &61.6(+1.4) &46.7(+1.2) &25.7(+0.0) &46.5(+0.9) &57.3(+2.4) \\
                            & FRS~\cite{zhixing2021frs}     &43.7(+1.7) &62.2(+2.0) &47.6(+2.1) &25.7(+0.0) &46.8(+1.2) &58.1(+3.2) \\
                            & FGFI\cite{wang2019fgfi}    &44.1(+2.1) &62.6(+2.4) &47.9(+2.4) &26.3(+0.6) &47.3(+1.7) &58.5(+3.6) \\
                            & FKD\cite{zhang2021improve}     &43.6(+1.6) &62.0(+1.8) &47.1(+1.6) &25.9(+0.2) &47.0(+1.4) &58.1(+3.2) \\
                            & FGD\cite{yang2021fgd}     &44.1(+2.1) &62.4(+2.2) &47.9(+2.4) &26.8(+1.1) &47.2(+1.6) &58.5(+3.6) \\
                            & Ours    &45.4(+\textbf{3.4}) &63.9(+\textbf{3.7}) &49.0(+\textbf{3.5}) &26.9(\textbf{+1.2}) &49.2(+\textbf{3.6}) &59.7(+\textbf{4.8}) \\
        \hline
    \end{tabularx}
    \end{threeparttable}
    \captionsetup{font={footnotesize}}
  \caption{A comparison between our \ourmethod~with other state-of-the-art distillation methods on COCO \textit{mini-val} set. All models are trained locally. We set hyper-parameters for competing methods following their paper or open-sourced code bases.}\label{tbl:big-compare}
\end{table*}

%% file: tabels/small_backbone.tex
\begin{table*}[hb!]
\renewcommand\arraystretch{1.2}
\scriptsize \setlength{\tabcolsep}{4.5pt}
  \centering
  \begin{threeparttable}
    \begin{tabularx}{\textwidth}{X|X|XXXXXX}
        \hline  
        \rowcolor{white}
        Detector & Setting & AP & AP$_{50}$ & AP$_{75}$ & AP$_{S}$ & AP$_{M}$ & AP$_{L}$ \\
        \hline 
        \multirow{4}{*}{FCOS} & \cellcolor{gray!25}Teacher  &\cellcolor{gray!25}43.1 &\cellcolor{gray!25}62.4 &\cellcolor{gray!25}46.6 &\cellcolor{gray!25}25.5 &\cellcolor{gray!25}47.1 &\cellcolor{gray!25}54.7 \\
        & \cellcolor{gray!25}Student  &\cellcolor{gray!25}32.8 &\cellcolor{gray!25}51.3 &\cellcolor{gray!25}34.5 &\cellcolor{gray!25}18.4 &\cellcolor{gray!25}35.4 &\cellcolor{gray!25}42.6 \\
        & FGD     &34.7(+1.9) &53.0(+1.7) &36.8(+2.3) &19.8(+1.4) &36.8(+1.4) &44.9(+2.3) \\
        & Ours    &37.3(+\textbf{4.5}) &55.6(+\textbf{4.3}) &39.8(+\textbf{5.3}) &20.5(+\textbf{2.1}) &40.3(+\textbf{4.9}) &49.9(+\textbf{7.3}) \\
        \hline
        \multirow{4}{*}{ATSS} & \cellcolor{gray!25}Teacher  &\cellcolor{gray!25}45.5 &\cellcolor{gray!25}63.9 &\cellcolor{gray!25}49.7 &\cellcolor{gray!25}28.7 &\cellcolor{gray!25}50.1 &\cellcolor{gray!25}57.8 \\
        & \cellcolor{gray!25}Student  &\cellcolor{gray!25}33.5 &\cellcolor{gray!25}50.1 &\cellcolor{gray!25}36.0 &\cellcolor{gray!25}18.7 &\cellcolor{gray!25}36.2 &\cellcolor{gray!25}43.6 \\
        & FGD     &35.8(+2.3) &52.6(+2.5) &38.8(+2.8) &20.6(+1.9) &38.4(+2.2) &46.2(+2.6) \\
        & Ours    &38.3(+\textbf{4.8}) &55.1(+\textbf{5.0}) &41.7(+\textbf{5.7}) &21.3(+\textbf{2.6}) &41.6(+\textbf{5.4}) &51.6(+\textbf{8.0}) \\
        \hline
    \end{tabularx}
    \end{threeparttable}
    \captionsetup{font={footnotesize}}
  \caption{Distillation results on COCO \textit{mini-val} using MobileNetV2 as the student backbone.}
 \label{tbl:lightweighted}
\end{table*}

%% file: tabels/crowdhuman.tex
\begin{SCtable*}[][h]
  \renewcommand\arraystretch{1.2}
  \scriptsize \setlength{\tabcolsep}{4.5pt}
  \begin{threeparttable}
    \begin{tabularx}{0.6\textwidth}{X|XXX}
      \hline
      ~Setting & ~MR $\downarrow$ & AP $\uparrow$ &  JI $\uparrow$\\
      \hline
      \cellcolor{gray!25}Teacher & \cellcolor{gray!25}~41.4 & \cellcolor{gray!25}90.2 & \cellcolor{gray!25}81.4 \\ 
      \cellcolor{gray!25}Student & \cellcolor{gray!25}~46.0 & \cellcolor{gray!25}88.0 & \cellcolor{gray!25}79.0 \\ 
      FKD~\cite{} & ~44.3(-1.7)~  & 89.1(+1.1)~ & 80.0(+1.0) \\
      DeFeat~\cite{guo2021defeat} & ~44.2(-1.8)~  & 89.1(+1.1)~ & 79.9(+0.9) \\
      FRS~\cite{zhixing2021frs} & ~44.1(-1.9)~  & 89.2(+1.2)~ & 80.3(+1.3) \\
      FGFI~\cite{wang2019fgfi} & ~43.8(-2.2)~  & 89.2(+1.2)~ & 80.3(+1.3) \\
      FGD~\cite{yang2021fgd} & ~43.1(-2.9)~  & 89.3(+1.3)~ & 80.4(+1.4) \\
      \hline
      Ours & ~42.8(-\textbf{3.2})~  & 90.0(+\textbf{2.0})~ & 80.7(+\textbf{1.7}) \\
      \hline
    \end{tabularx}
    \end{threeparttable}
  \captionsetup{font={footnotesize}}
  \caption{A comparison between our \ourmethod~with other state-of-the-art distillation methods on CrowdHuman \textit{validation} set using DDOD as object detector. }
 \label{tbl:crowdhuman}
\end{SCtable*}

%% file: tabels/selfdistill.tex
\begin{table*}[h]
\renewcommand\arraystretch{1.2}
\scriptsize \setlength{\tabcolsep}{4.5pt}
  \centering
  \begin{threeparttable}
    \begin{tabularx}{\textwidth}{X|X|XXXXXX}
        \hline
        \rowcolor{white}
        Detector & Setting & AP & AP$_{50}$ & AP$_{75}$ & AP$_{S}$ & AP$_{M}$ & AP$_{L}$ \\
        \hline  
        \multirow{3}{*}{FCOS} &\cellcolor{gray!25}S \& T &\cellcolor{gray!25}38.2 &\cellcolor{gray!25}57.9 &\cellcolor{gray!25}40.5 &\cellcolor{gray!25}23.1 &\cellcolor{gray!25}41.3 &\cellcolor{gray!25}49.4 \\
        & FGD     &39.0(+0.8) &58.6(+0.7) &41.4(+0.9) &23.7(+0.6) &42.1(+0.8) &50.6(+1.2) \\
        & Ours    &39.5(+1.3) &59.2(+1.3) &41.9(+1.4) &24.4(+1.3) &42.8(+1.5) &50.6(+1.2) \\
        \hline
        \multirow{3}{*}{ATSS} & \cellcolor{gray!25}S \& T  &\cellcolor{gray!25}39.6 &\cellcolor{gray!25}57.6 &\cellcolor{gray!25}43.2 &\cellcolor{gray!25}23.0 &\cellcolor{gray!25}42.9 &\cellcolor{gray!25}51.2 \\
        & FGD     &40.2(+0.6) &58.6(+1.0) &43.6(+1.4) &23.3(+0.3) &43.7(+0.8) &52.3(+1.1) \\
        & Ours    &40.7(+1.1) &58.9(+1.3) &44.2(+2.0) &24.0(+0.9) &44.2(+1.3) &52.9(+1.7) \\
        \hline
    \end{tabularx}
    \end{threeparttable}
    \captionsetup{font={footnotesize}}
  \caption{Self-distillation performance on COCO \textit{mini-val}. ResNet-50 is adopted as teacher and student backbone, which are both trained for 1$\times$ schedule.}
 \label{tbl:selfdistill}
\end{table*}

%% file: tabels/foreground.tex
\begin{table*}[t!]
\renewcommand\arraystretch{1.2}
\scriptsize \setlength{\tabcolsep}{4.5pt}
  \centering
  \begin{threeparttable}
    \begin{tabularx}{\textwidth}{X|XXXXXX}
        \hline  
        \rowcolor{white}
        Setting & AP & AP$_{50}$ & AP$_{75}$ & AP$_{S}$ & AP$_{M}$ & AP$_{L}$ \\
        \hline 
         \cellcolor{gray!25}Teacher  &\cellcolor{gray!25}45.5 &\cellcolor{gray!25}63.9 &\cellcolor{gray!25}49.7 &\cellcolor{gray!25}28.7 &\cellcolor{gray!25}50.1 &\cellcolor{gray!25}57.8  \\
         \cellcolor{gray!25}Student  &\cellcolor{gray!25}39.6 &\cellcolor{gray!25}57.6 &\cellcolor{gray!25}43.2 &\cellcolor{gray!25}23.0 &\cellcolor{gray!25}42.9 &\cellcolor{gray!25}51.2 \\
         Box  &43.3 (+3.7) &61.4(+3.8) &47.2(+4.0) &25.9(+2.9) &47.6(+4.7) &56.4(+5.2) \\
         BoxGauss &43.7(+4.1) &61.9(+4.3) &47.6(+4.4) &26.7(+3.7) &47.8(+4.9) &56.6(+5.4) \\
         Centre &43.1(+3.5) &61.0(+3.4) &46.9(+3.7) &25.9(+2.9) &47.3(+4.4) &56.1(+4.9) \\
         Quality  &43.8(+4.2) &61.8(+4.2) &47.8(+4.6) &25.7(+2.7) &48.2(+5.3) &56.8(+5.6) \\
         TopkEq &43.9(+4.3) &62.0(+4.4) &47.7(+4.5) &27.1(+\textbf{4.1}) &48.0(+5.1) &56.8(+5.6) \\
         KDE     &44.0(+4.4) &62.1(+4.5) &47.8(+4.6) &26.3(+3.3) &48.5(+5.6) &56.8(+5.6) \\
        \hline 
        Ours    &44.2(+\textbf{4.6}) &62.3(+\textbf{4.7}) &48.3(+\textbf{5.1}) &26.5(+3.5) &48.6(+\textbf{5.7}) &57.1(+\textbf{5.9}) \\
        \hline
    \end{tabularx}
    \end{threeparttable}
    \captionsetup{font={footnotesize}}
  \caption{Ablation study on different foreground distillation strategies on COCO \textit{mini-val} seet using ATSS as object detector.}
 \label{tbl:foreground}
\end{table*}

%% file: tabels/hyperparameter.tex
\begin{table*}[ht!]
    \renewcommand\arraystretch{1.2}
  \scriptsize \setlength{\tabcolsep}{4.5pt}
    \begin{subtable}[h]{\textwidth}
        \centering
        \begin{threeparttable}
        \begin{tabularx}{\textwidth}{C|CCCCCCC}
        \hline
        K & 1 & 5 & 9 & 15 & 30 & 45 & 60 \\
        \hline
        AP & 43.2 & 43.5 & 43.6 & 43.9 & \textbf{44.2} & 44.0 & 43.9 \\
        \hline
       \end{tabularx}
       \captionsetup{font={scriptsize}}
       \caption{Ablation study on different K in the top-K operation using ATSS as detector.}
       \end{threeparttable}
    \end{subtable}
    \begin{subtable}[h]{\textwidth}
        \centering
        \begin{threeparttable}
        \begin{tabularx}{\textwidth}{C|CCCCCCC}
        \hline
        $\alpha$ & 0.005 & 0.01 & 0.03 & 0.05 & 0.07 & 0.1 & 0.2 \\
        \hline
        FCOS & 41.7 & 42.0 & \textbf{42.5} & \textbf{42.5} & 42.4 & 42.2 & 41.8 \\
        \hline
        ATSS & 42.9 & 43.2 & 43.7 & 43.9 & \textbf{44.2} & 44.1 & 43.2 \\
        \hline
       \end{tabularx}
       \captionsetup{font={scriptsize}}
        \caption{Ablation study on distillation loss magnitude $\alpha$ using FCOS and ATSS.}
       \end{threeparttable}
     \end{subtable}
     \captionsetup{font={footnotesize}}
     \caption{Hyper-parameter ablation studies on COCO \textit{mini-val}.}
     \label{tbl:hyperparameter}
\end{table*}

%% file: figures/tex/qulitative.tex
\begin{figure*}[!t]
    \centering
    \includegraphics[width=\textwidth]{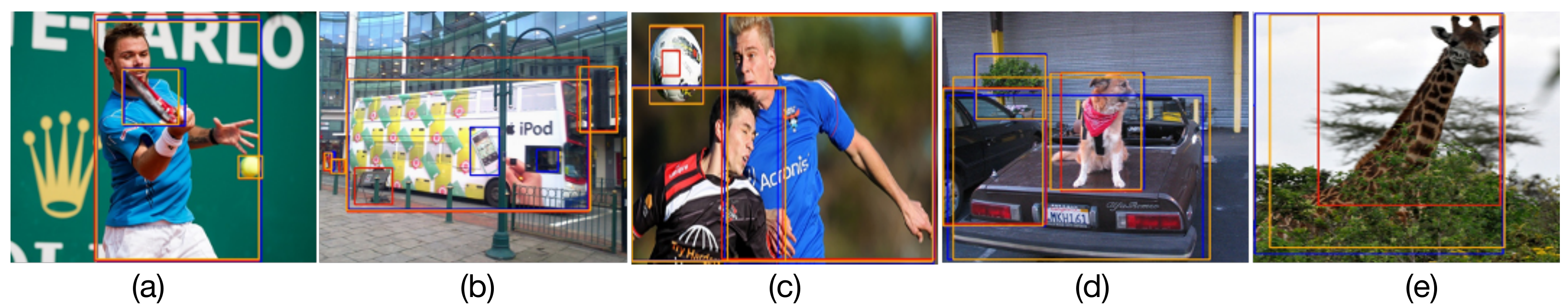}
    \captionsetup{font={footnotesize}}
    \caption{Visualisation of the detection results on COCO \textit{mini-val} set using ATSS as detector and \ourmethod~for distillation. GTs are shown in \textcolor{VisBlue}{blue}; plain student detections are shown in \textcolor{VisRed}{red}; distilled student predictions are shown in \textcolor{VisOrange}{orange}.}\label{fig:qualitative}
\end{figure*}

%% file: tabels/distill_branch.tex
\begin{table*}[h]
\renewcommand\arraystretch{1.2}
\scriptsize \setlength{\tabcolsep}{4.5pt}
  \centering
  \begin{threeparttable}
    \begin{tabularx}{\textwidth}{lll|XXXXXX}
        \hline   
        \rowcolor{white}
        neck & cls & reg & AP & AP$_{50}$ & AP$_{75}$ & AP$_{S}$ & AP$_{M}$ & AP$_{L}$ \\
        \hline 
          -  & - & - &39.6 &57.6 &43.2 &23.0 &42.9 & 51.2 \\
         \hline    
         \checkmark &  &  &43.6(+4.0) &61.8(+4.2) &47.5(+4.3) &26.1(+\textbf{3.1}) &47.8(+4.9) &56.8(+5.6) \\
         \hline
         & \checkmark &  &43.8(+4.2) &62.1(+4.5) &47.5(+4.3) &26.5(3.5) &48.0(+5.1) &56.8(+5.6) \\
         &  & \checkmark &41.7(+2.1) &60.2(+2.6) &45.2(+2.0) &25.3(+2.3) &45.4(+2.5) &53.9(+2.7) \\
         & \checkmark & \checkmark &44.2(+\textbf{4.6}) &62.3(+\textbf{4.7}) &48.3(+\textbf{5.1}) &26.5(+3.5) &48.6(+\textbf{5.7}) &57.1(+\textbf{5.9}) \\
        \hline
    \end{tabularx}
    \end{threeparttable}
    \captionsetup{font={footnotesize}}
  \caption{Comparison between different distillation branches.}
 \label{tbl:branch}
\end{table*}

%% file: sections/conclusion.tex
\section{Conclusion}
In this work, we highlight the need to focus distillation on features of the teacher that are responsible for high-scoring predictions. We find that these \disregion~constitute only a small fraction of all features within the boundaries of the ground-truth bounding box. We use this observation to design a novel distillation technique---\ourmethod---that amplifies the distillation signal from these features. We use an adaptive Gaussian distribution to smoothly aggregate those top locations to further enhance performance. Our approach can significantly improve state-of-the-art detectors on COCO and CrowdHuman, outperforming many existing KD methods. In future, we could investigate the applicability of high-quality regions to two-stage and transformer models for detection.

\section*{Acknowledgements}
The authors would like to thank Joe Mellor, Kaihong Wang, and Zehui Chen for their useful comments and suggestions. This work was supported by a PhD studentship provided by the School of Engineering, University of Edinburgh as well as the EPSRC Centre for Doctoral Training in Robotics and Autonomous Systems (Grant No. EP/S515061/1) and SeeByte Ltd, Edinburgh, UK.

%% file: sections/supp.tex
\section{Additional Experiments}
\subsection{Weak to strong distillation}
We conduct experiments to show that our method is also effective when distilling a weak teacher into a stronger model. Specifically, we use ATSS~\cite{zhang2020bridging} as the detector and use a ResNet-50 based model as teacher with a ResNet-101 based model as student. The results are shown in Table~\ref{tbl:weak}. We observe that the student AP is improved to 44.2 from 41.4, even surpassing the teacher performance. A possible reason for this is that the student learns from the key predictive regions where it should focus on. Then, empowered by its stronger backbone, the student is able to surpass the teacher model.
\vspace{-0.3cm}

\begin{table*}[h!]
\renewcommand\arraystretch{1.2}
\scriptsize \setlength{\tabcolsep}{4.5pt}
  \centering
  \begin{threeparttable}
    \begin{tabularx}{\textwidth}{l|XXX|XXX}
    \hline
    Model &AP & AP$_{50}$ & AP$_{75}$ & AP$_{S}$  & AP$_{M}$ & AP$_{L}$   \\
    \hline
    R50 (ms, 3$\times$) & 43.6 & 61.8 & 47.4 & 28.5 & 48.1 & 54.3 \\
    R101 (1$\times$)& 41.4 & 59.8 & 45.2 & 24.2 & 45.8 & 53.8 \\
    R50$\rightarrow$R101 (1$\times$) & 44.2 & 62.7 & 47.9 & 28.8 & 48.7 & 55.7 \\
    \hline
    \end{tabularx}
    \end{threeparttable}
  \captionsetup{font={footnotesize}}
  \caption{Distillation results using our PGD on ATSS detector. Training settings are inherited from the paper: teacher is trained for 3$\times$ schedule with multi-scale input; Student and KD models are trained for 1$\times$ schedule with single-scale input.}
 \label{tbl:weak}
\end{table*}

\vspace{-10mm}
\subsection{Improvement Discussion.} 
\begin{figure*}[!h]
    \centering
    \includegraphics[width=\textwidth]{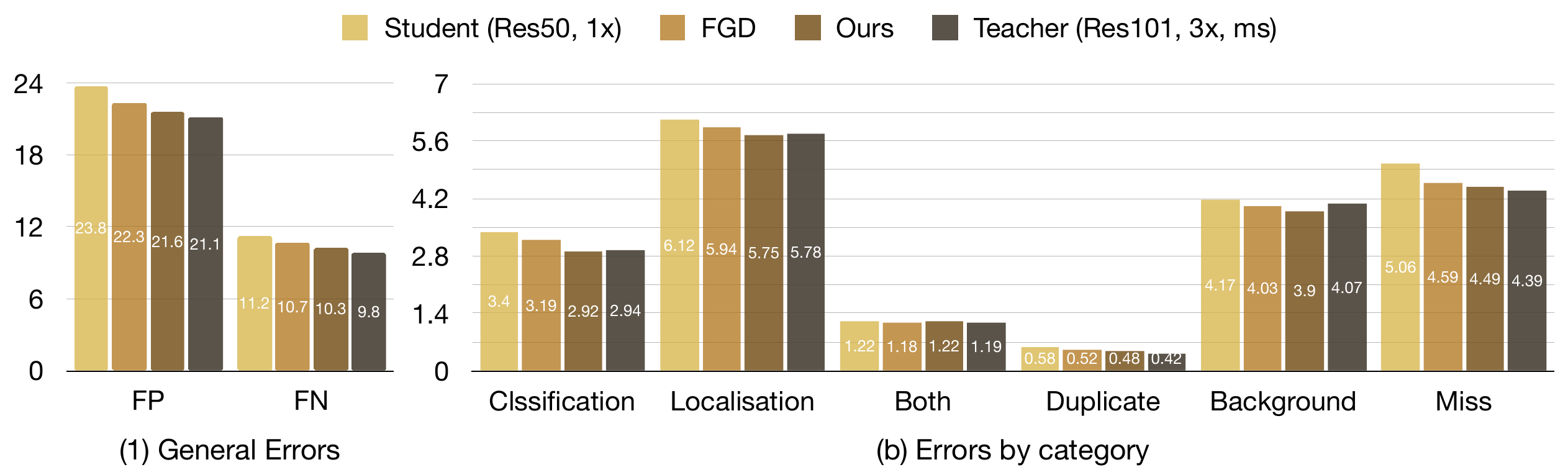}
    \captionsetup{font={footnotesize}}
    \caption{TIDE error analysis on COCO \textit{mini-val} using ATSS as the object detector.}
    \label{fig:tide}
\end{figure*}

We used the TIDE~\cite{tide-eccv2020} toolkit to analyse the performance improvement after KD and compare our approach with the baseline FGD. We use ATSS as the object detector and present the COCO \textit{mini-val} evaluation analysis under 0.5 IoU threshold in Fig.~\ref{fig:tide}. For the general errors, we observe that our \ourmethod~can significantly improve both false negatives and false positives over the baseline FGD. When we look into different categories of errors, we find \ourmethod~can improve most error cases, including \textit{Classification}, \textit{Localisation}, \textit{Duplicate}, \textit{Background}, and \textit{Missing} errors. Moreover, the student model improved by our method makes fewer~\textit{Localisation} and \textit{Background} errors than the teacher model; our method can improve a student model's localisation ability and the ability to ignore background noise. However, it does not lead to a reduction in \textit{Both} errors, which happens when classification and localisation are both incorrect. However, as this error is similar between the teacher and plain student models, we attribute this problem to training randomness.

We also used the recently proposed LRP~\cite{lrp} metrics to compare our method with the baseline FGD. The results are shown in Table~\ref{tbl:lrp}. Our method improves the student detector's LRP performance in both classification and localisation comparing with the FGD, illustrating the effectiveness of our approach in improving the student’s detection ability.

\begin{table*}[h!]
\renewcommand\arraystretch{1.2}
\scriptsize \setlength{\tabcolsep}{4.5pt}
  \centering
  \begin{threeparttable}
    \begin{tabularx}{\textwidth}{l|X|XXXX}
        \hline   
        \rowcolor{white}
        Setting & AP & LRP & LRP$_{loc}$ & LRP$_{FP}$ & LRP$_{FN}$ \\
        \hline 
        Teacher (R101, 3$\times$, ms) & 45.5 & 63.5 & 14.1 & 26.2 & 41.0 \\
        Student (R50, 1$\times$)& 39.4 & 68.5 & 15.3 & 30.5 & 46.5 \\
        FGD & 42.6 & 66.0 & 14.5 & 28.0 & 44.1 \\
        PGD (Ours) & 44.2 & 64.6 & 14.2 & 26.1 & 43.0 \\
        \hline
    \end{tabularx}
    \end{threeparttable}
    \captionsetup{font={footnotesize}}
  \caption{LRP analysis on COCO mini-val using ATSS as the object detector}
 \label{tbl:lrp}
\end{table*}

\subsection{COCO \textit{Test-Dev} results}
Knowledge distillation aims to equip a lightweight model with a strong generalisation capability. With this in mind, we compare a detector produced using our method with the most state-of-the-art dense object detectors on the COCO \textit{test-dev} set. We used DDOD~\cite{chen2021ddod} as the object detector; ResNet-101 is used as the student backbone, and Res2Net-DCN~\cite{gao2019res2net,dai2017deformable} is used as the teacher backbone. We train the model using COCO \textit{train-2017} set with 2$\times$ training schedule (24 epochs) and multi-scale input. The results are presented in Tab~\ref{tbl:testdev}, from which we can see that the model trained using our method achieves the highest AP, suggestion that our approach can indeed improve a detector's generalisation ability.
\begin{table*}[h!]
\renewcommand\arraystretch{1.2}
\scriptsize \setlength{\tabcolsep}{4.5pt}
  \centering
  \begin{threeparttable}
    \begin{tabularx}{\textwidth}{l|XXXXXX}
        \hline   
        \rowcolor{white}
        Setting & AP & AP$_{50}$ & AP$_{75}$ & AP$_{S}$ & AP$_{M}$ & AP$_{L}$ \\
        \hline 
         RetinaNet~\cite{lin2017focal} &39.1 &59.1 &42.3 &21.8 &42.7 &50.2 \\
         FCOS~\cite{tian2019fcos}  &41.5 & 60.7 & 45.0 & 24.4 & 44.8 & 51.6 \\
         FreeAnchor~\cite{NEURIPS2019_43ec517d} &43.1 &62.2 &46.4 &24.5 &46.1 &54.8 \\
         SAPD~\cite{zhu2019soft} & 43.5 & 63.6 & 46.5 & 24.9 & 46.8 & 54.6 \\
         ATSS~\cite{zhang2020bridging} &43.6 &62.1 &47.4 &26.1 &47.0 &53.6 \\
         AutoAssign~\cite{zhu2020autoassign} &44.5 &64.3 &48.4 &25.9 &47.4 &55.0 \\
         PAA~\cite{kim2020probabilistic}  & 44.8 & 63.3 & 48.7 & 26.5 & 48.8 & 56.3 \\
         GFL~\cite{li2020generalized} & 45.0 & 63.7 & 48.9 & 27.2 & 48.8 & 54.5 \\
         IQDet~\cite{Ma2021iqdet} &45.1 &63.4 &49.3 &26.7 &48.5 &56.6 \\ 
         OTA~\cite{Ge2021OTA} &45.3 &63.5 &49.3 &26.9 &48.8 &56.1 \\
         GFLv2~\cite{Li_2021GFLv2} &46.0 &64.1 &50.2 &27.6 &49.6 &56.5 \\
         VFNet~\cite{Zhang_2021VFNet} &46.0 &64.2 &50.0 &27.5 &49.4 & 56.9\\
         RepPointsV2~\cite{chen2020reppointsv2} &46.0 &65.3 &49.5 &27.4 &48.9 &57.3\\
         DDOD~\cite{chen2021ddod}  & 46.7 & 65.3 & 51.1 & 28.2 & 49.9 & 57.9 \\
        \hline 
        Ours    & \textbf{48.2} & \textbf{66.9} & \textbf{52.5} & \textbf{30.1} & \textbf{51.6} & \textbf{58.5} \\
        \hline
    \end{tabularx}
    \end{threeparttable}
    \captionsetup{font={footnotesize}}
  \caption{Comparison with state-of-the-art dense object detectors on COCO \textit{test-dev}. All models are trained for 2$\times$ (24 epochs) with ResNet-101 as backbone.}
 \label{tbl:testdev}
\end{table*}

%% file: main.bbl
\begin{thebibliography}{10}
\providecommand{\url}[1]{\texttt{#1}}
\providecommand{\urlprefix}{URL }
\providecommand{\doi}[1]{https://doi.org/#1}

\bibitem{ba2014do}
Ba, L.J., Caruana, R.: Do deep nets really need to be deep? In: NeurIPS (2014)

\bibitem{tide-eccv2020}
Bolya, D., Foley, S., Hays, J., Hoffman, J.: {TIDE: A General Toolbox for
  Identifying Object Detection Errors}. In: ECCV (2020)

\bibitem{cai2018cascade}
Cai, Z., Vasconcelos, N.: {Cascade R-CNN: Delving into high quality object
  detection}. In: CVPR (2018)

\bibitem{carion2020end}
Carion, N., Massa, F., Synnaeve, G., Usunier, N., Kirillov, A., Zagoruyko, S.:
  {End-to-End Object Detection with Transformers}. In: ECCV (2020)

\bibitem{chen2017learning}
Chen, G., Choi, W., Yu, X., Han, T., Chandraker, M.: {Learning efficient object
  detection models with knowledge distillation}. In: NeurIPS (2017)

\bibitem{chen2019mmdetection}
Chen, K., et~al.: Mmdetection: Open mmlab detection toolbox and benchmark. In:
  arXiv preprint arXiv:1906.07155 (2019)

\bibitem{chen2020reppointsv2}
Chen, Y., Zhang, Z., Cao, Y., Wang, L., Lin, S., Hu, H.: Reppoints v2:
  Verification meets regression for object detection. In: NeurIPS (2020)

\bibitem{chen2021ddod}
Chen, Z., Yang, C., Li, Q., Zhao, F., Zha, Z.J., Wu, F.: Disentangle your dense
  object detector. In: ACM MM (2021)

\bibitem{Chu2020crowddet}
Chu, X., Zheng, A., Zhang, X., Sun, J.: {Detection in Crowded Scenes: One
  Proposal, Multiple Predictions}. In: CVPR (2020)

\bibitem{dai2017deformable}
Dai, J., Qi, H., Xiong, Y., Li, Y., Zhang, G., Hu, H., Wei, Y.: Deformable
  convolutional networks. In: ICCV (2017)

\bibitem{dai2021gid}
Dai, X., Jiang, Z., Wu, Z., Bao, Y., Wang, Z., Liu, S., Zhou, E.: General
  instance distillation for object detection. In: CVPR (2021)

\bibitem{gao2019res2net}
Gao, S., Cheng, M.M., Zhao, K., Zhang, X.Y., Yang, M.H., Torr, P.H.: Res2net: A
  new multi-scale backbone architecture  (2021).
  \doi{10.1109/TPAMI.2019.2938758}

\bibitem{Ge2021OTA}
Ge, Z., Liu, S., Li, Z., Yoshie, O., Sun, J.: {OTA: Optimal Transport
  Assignment for Object Detection}. In: CVPR (2021)

\bibitem{guo2021defeat}
Guo, J., Han, K., Wang, Y., Wu, H., Chen, X., Xu, C., Xu, C.: Distilling object
  detectors via decoupled features. In: CVPR (2021)

\bibitem{he2019rethinking}
He, K., Girshick, R., Doll{\'a}r, P.: {Rethinking Imagenet Pre-training}. In:
  CVPR (2019)

\bibitem{he2017mask}
He, K., Gkioxari, G., Doll{\'a}r, P., Girshick, R.: {Mask r-cnn}. In: ICCV
  (2017)

\bibitem{he2016deep}
He, K., Zhang, X., Ren, S., Sun, J.: {Deep residual learning for image
  recognition}. In: CVPR (2016)

\bibitem{hinton2015distilling}
Hinton, G., Vinyals, O., Dean, J., et~al.: Distilling the knowledge in a neural
  network. In: NeurIPS 2014 Deep Learning Workshop (2014)

\bibitem{kang2021instance}
Kang, Z., Zhang, P., Zhang, X., Sun, J., Zheng, N.: {Instance-Conditional
  Knowledge Distillation for Object Detection}. In: NeurIPS (2021)

\bibitem{kim2020probabilistic}
Kim, K., Lee, H.S.: Probabilistic anchor assignment with iou prediction for
  object detection. In: ECCV (2020)

\bibitem{Li_2021GFLv2}
Li, X., Wang, W., Hu, X., Li, J., Tang, J., Yang, J.: {Generalized Focal Loss
  V2: Learning Reliable Localization Quality Estimation for Dense Object
  Detection}. In: CVPR (2021)

\bibitem{li2020generalized}
Li, X., Wang, W., Wu, L., Chen, S., Hu, X., Li, J., Tang, J., Yang, J.:
  Generalized focal loss: Learning qualified and distributed bounding boxes for
  dense object detection. In: NeurIPS (2020)

\bibitem{li2019scale}
Li, Y., Chen, Y., Wang, N., Zhang, Z.: {Scale-aware trident networks for object
  detection}. In: ICCV (2019)

\bibitem{li2017light}
Li, Z., Peng, C., Yu, G., Zhang, X., Deng, Y., Sun, J.: Light-head r-cnn: In
  defense of two-stage object detector. In: arXiv preprint arXiv:1711.07264
  (2017)

\bibitem{lin2017focal}
Lin, T.Y., Goyal, P., Girshick, R., He, K., Doll{\'a}r, P.: {Focal loss for
  dense object detection}. In: ICCV (2017)

\bibitem{lin2014microsoft}
Lin, T.Y., Maire, M., Belongie, S., Hays, J., Perona, P., Ramanan, D.,
  Doll{\'a}r, P., Zitnick, C.L.: {Microsoft {COCO}: Common objects in context}.
  In: ECCV (2014)

\bibitem{Ma2021iqdet}
Ma, Y., Liu, S., Li, Z., Sun, J.: {IQDet: Instance-Wise Quality Distribution
  Sampling for Object Detection}. In: CVPR (2021)

\bibitem{Nguyen2022lad}
Nguyen, C.H., Nguyen, T.C., Tang, T.N., Phan, N.L.: Improving object detection
  by label assignment distillation. In: WACV (2022)

\bibitem{oksuz2020ranking}
Oksuz, K., Cam, B.C., Akbas, E., Kalkan, S.: A ranking-based, balanced loss
  function unifying classification and localisation in object detection.
  Advances in Neural Information Processing Systems  \textbf{33},  15534--15545
  (2020)

\bibitem{lrp}
Oksuz, K., Cam, B.C., Kalkan, S., Akbas, E.: One metric to measure them all:
  Localisation recall precision (lrp) for evaluating visual detection tasks.
  TPAMI  (2021). \doi{10.1109/TPAMI.2021.3130188}

\bibitem{paszke2019pytorch}
Paszke, A., Gross, S., Massa, F., Lerer, A., Bradbury, J., Chanan, G., Killeen,
  T., Lin, Z., Gimelshein, N., Antiga, L., et~al.: {Pytorch: An imperative
  style, high-performance deep learning library}. In: NeurIPS (2019)

\bibitem{qin2019thundernet}
Qin, Z., Li, Z., Zhang, Z., Bao, Y., Yu, G., Peng, Y., Sun, J.: Thundernet:
  Towards real-time generic object detection on mobile devices. In: ICCV (2019)

\bibitem{RedmonDiGiFa16}
Redmon, J., Divvala, S., Girshick, R., Farhadi, A.: {You only look once:
  Unified, real-time object detection}. In: CVPR (2016)

\bibitem{redmon2017yolo9000}
Redmon, J., Farhadi, A.: {YOLO9000: better, faster, stronger}. In: CVPR (2017)

\bibitem{ren2015faster}
Ren, S., He, K., Girshick, R., Sun, J.: {Faster R-CNN: Towards real-time object
  detection with region proposal networks}. In: NeurIPS (2015)

\bibitem{sandler2018mobilenetv2}
Sandler, M., Howard, A., Zhu, M., Zhmoginov, A., Chen, L.C.: {Mobilenetv2:
  Inverted Residuals and Linear Bottlenecks}. In: CVPR (2018)

\bibitem{shao2018crowdhuman}
Shao, S., Zhao, Z., Li, B., Xiao, T., Yu, G., Zhang, X., Sun, J.: Crowdhuman: A
  benchmark for detecting human in a crowd. In: arXiv preprint arXiv:1805.00123
  (2018)

\bibitem{sun2020tadf}
Sun, R., Tang, F., Zhang, X., Xiong, H., Tian, Q.: Distilling object detectors
  with task adaptive regularization. In: arXiv preprint arXiv:2006.13108 (2020)

\bibitem{tian2019fcos}
Tian, Z., Shen, C., Chen, H., He, T.: {FCOS: Fully convolutional one-stage
  object detection}. In: ICCV (2019)

\bibitem{wang2021end}
Wang, J., Song, L., Li, Z., Sun, H., Sun, J., Zheng, N.: {End-to-end object
  detection with fully convolutional network}. In: CVPR (2021)

\bibitem{wang2018pelee}
Wang, R.J., Li, X., Ling, C.X.: Pelee: A real-time object detection system on
  mobile devices. In: NeurIPS (2018)

\bibitem{wang2019fgfi}
Wang, T., Yuan, L., Zhang, X., Feng, J.: Distilling object detectors with
  fine-grained feature imitation. In: CVPR (2019)

\bibitem{yang2021fgd}
Yang, Z., Li, Z., Jiang, X., Gong, Y., Yuan, Z., Zhao, D., Yuan, C.: Focal and
  global knowledge distillation for detectors. In: arXiv preprint
  arXiv:2111.11837 (2021)

\bibitem{yao2021gdetkd}
Yao, L., Pi, R., Xu, H., Zhang, W., Li, Z., Zhang, T.: G-detkd: Towards general
  distillation framework for object detectors via contrastive and
  semantic-guided feature imitation. In: ICCV (2021)

\bibitem{Zhang_2021VFNet}
Zhang, H., Wang, Y., Dayoub, F., Sunderhauf, N.: {VarifocalNet: An IoU-Aware
  Dense Object Detector}. In: CVPR) (2021)

\bibitem{zhang2021improve}
Zhang, L., Ma, K.: {Improve Object Detection with Feature-based Knowledge
  Distillation: Towards Accurate and Efficient Detectors}. In: ICLR (2021)

\bibitem{zhang2020bridging}
Zhang, S., Chi, C., Yao, Y., Lei, Z., Li, S.Z.: {Bridging the gap between
  anchor-based and anchor-free detection via adaptive training sample
  selection}. In: CVPR (2020)

\bibitem{NEURIPS2019_43ec517d}
Zhang, X., Wan, F., Liu, C., Ji, R., Ye, Q.: Freeanchor: Learning to match
  anchors for visual object detection  (2019)

\bibitem{zheng2020distance}
Zheng, Z., Wang, P., Liu, W., Li, J., Ye, R., Ren, D.: Distance-iou loss:
  Faster and better learning for bounding box regression. In: AAAI (2020)

\bibitem{zheng2021ld}
Zheng, Z., Ye, R., Wang, P., Wang, J., Ren, D., Zuo, W.: Localization
  distillation for object detection. In: arXiv preprint arXiv:2102.12252 (2021)

\bibitem{zhixing2021frs}
Zhixing, D., Zhang, R., Chang, M., Liu, S., Chen, T., Chen, Y., et~al.:
  Distilling object detectors with feature richness. In: NeurIPS (2021)

\bibitem{zhu2020autoassign}
Zhu, B., Wang, J., Jiang, Z., Zong, F., Liu, S., Li, Z., Sun, J.: Autoassign:
  Differentiable label assignment for dense object detection. In: arXiv
  preprint arXiv:2007.03496 (2020)

\bibitem{zhu2019soft}
Zhu, C., Chen, F., Shen, Z., Savvides, M.: {Soft anchor-point object detection}
   (2020)

\end{thebibliography}
